\documentclass[conference]{IEEEtran}
\IEEEoverridecommandlockouts

\usepackage{cite}
\usepackage{amsmath,amssymb,amsfonts}
\usepackage{graphicx}
\usepackage{textcomp}
\usepackage{xcolor}
\def\BibTeX{{\rm B\kern-.05em{\sc i\kern-.025em b}\kern-.08em
    T\kern-.1667em\lower.7ex\hbox{E}\kern-.125emX}}

\usepackage{placeins}  
\usepackage{caption}  
\usepackage{threeparttable}
\usepackage{multirow}
\usepackage{booktabs}
\usepackage{array}
\usepackage{makecell}
\usepackage{url}

\usepackage{algorithm}
\usepackage{algpseudocode}

\makeatletter
\newcommand{\removelatexerror}{\let\@latex@error\@gobble}
\makeatother

\begin{document}

\title{FedL2T: Personalized Federated Learning with Two-Teacher Distillation for Seizure Prediction\\
}

\makeatletter
\newcommand{\linebreakand}{\end{@IEEEauthorhalign}\hfill\mbox{}\par
\mbox{}\hfill\begin{@IEEEauthorhalign}}
\makeatother

\author{
\IEEEauthorblockN{\textsuperscript{} Jionghao Lou}
\IEEEauthorblockA{
    \textit{Key Laboratory of Smart Manufacturing} \\
    \textit{in Energy Chemical Process, Ministry of Education} \\
    \textit{East China University of Science and Technology} \\
    Shanghai, China \\
    jhlou@mail.ecust.edu.cn}
\and
\IEEEauthorblockN{\textsuperscript{} Jian Zhang}
\IEEEauthorblockA{
    \textit{School of Information Science and Engineering} \\
    \textit{East China University of Science and Technology} \\
    Shanghai, China \\
    zhjmaster@sina.com}
\linebreakand 
\IEEEauthorblockN{\textsuperscript{} Zhongmei Li}
\IEEEauthorblockA{ 
    \textit{Key Laboratory of Smart Manufacturing} \\
    \textit{in Energy Chemical Process, Ministry of Education} \\
    \textit{East China University of Science and Technology} \\
    Shanghai, China \\
    lizhongmei@sina.cn}
\and
\IEEEauthorblockN{\textsuperscript{} Lanlan Chen}
\IEEEauthorblockA{ 
    \textit{Key Laboratory of Smart Manufacturing} \\
    \textit{in Energy Chemical Process, Ministry of Education} \\
    \textit{East China University of Science and Technology} \\
    Shanghai, China \\
    llchen@ecust.edu.cn}
\linebreakand
\IEEEauthorblockN{\textsuperscript{} Enbo Feng*}
\IEEEauthorblockA{ 
    \textit{Key Laboratory of Smart Manufacturing} \\
    \textit{in Energy Chemical Process, Ministry of Education} \\
    \textit{East China University of Science and Technology} \\
    Shanghai, China \\
    enbo.feng@outlook.com}}


\maketitle

\begin{abstract}
The training of deep learning models in seizure prediction requires large amounts of Electroencephalogram (EEG) data. However, acquiring sufficient labeled EEG data is difficult due to annotation costs and privacy constraints. Federated Learning (FL) enables privacy-preserving collaborative training by sharing model updates instead of raw data. However, due to the inherent inter-patient variability in real-world scenarios, existing FL-based seizure prediction methods struggle to achieve robust performance under heterogeneous client settings. To address this challenge, we propose FedL2T, a personalized federated learning framework that leverages a novel two-teacher knowledge distillation strategy to generate superior personalized models for each client. Specifically, each client simultaneously learns from a globally aggregated model and a dynamically assigned peer model, promoting more direct and enriched knowledge exchange. To ensure reliable knowledge transfer, FedL2T employs an adaptive multi-level distillation strategy that aligns both prediction outputs and intermediate feature representations based on task confidence. In addition, a proximal regularization term is introduced to constrain personalized model updates, thereby enhancing training stability. Extensive experiments on two EEG datasets demonstrate that FedL2T consistently outperforms state-of-the-art FL methods, particularly under low-label conditions. Moreover, FedL2T exhibits rapid and stable convergence toward optimal performance, thereby reducing the number of communication rounds and associated overhead. These results underscore the potential of FedL2T as a reliable and personalized solution for seizure prediction in privacy-sensitive healthcare scenarios.

\end{abstract}


\section{Introduction}
Epilepsy is a chronic neurological disorder caused by abnormal synchronous discharges in the brain, affecting over 50 million people worldwide\cite{sunEmpoweringCrossPatientEpilepsy2024}. Electroencephalography (EEG), which records brain activity through electrodes placed on the scalp, has become the primary technique for epilepsy analysis due to its non-invasiveness and high temporal resolution. Clinical studies have shown that a pre-ictal state, occurring several minutes before a seizure, is distinguishable from both seizure (ictal) and normal (inter-ictal) states in EEG recordings \cite{segalPersonalizedPreictalEEG2025}. This distinction between pre-ictal and inter-ictal states makes seizure prediction possible.

In recent years, deep learning has demonstrated remarkable success in predicting seizures \cite{liSpatioTemporalSpectralHierarchicalGraph2022}. However, this success largely depends on the availability of large-scale datasets. However, collecting sufficient EEG data for seizure prediction remains challenging due to the infrequent nature of seizures and the fact that EEG acquisition is typically confined to clinical settings. The scarcity of data underscores the importance of cross-institutional data sharing. 

Nevertheless, privacy and ethical concerns, such as those mandated by the GDPR\cite{albrecht2016how}, severely restrict centralized data collection. Federated Learning (FL)\cite{McMahanMRHA17} provides a promising solution by enabling collaborative model training without sharing raw data. A representative approach is the classical FedAvg algorithm\cite{McMahanMRHA17}, which enables data owners to conduct local training and iteratively aggregate their models on a central server. However, directly substituting local models with the aggregated global model in each round may degrade performance, especially under non-independent and identically distributed (non-IID) data settings.
To mitigate this issue, personalized Federated Learning (pFL) \cite{zhang2025pfllib} allows each client to maintain a personalized model tailored to its local data, unlike traditional FL which uses a shared global model. For instance, Ditto\cite{liDittoFairRobust2021} introduces a proximal regularization term to stabilize local updates, while FedBN \cite{liFedBNFederatedLearning2020} and FedRep\cite{collinsExploitingSharedRepresentations2021} adopt personalized batch normalization and prediction layers, respectively, to better align feature representations across heterogeneous clients. In addition, knowledge distillation (KD) techniques\cite{moraKnowledgeDistillationFederated2024} have been employed to prevent local models from being overwritten by the global model. In KD-based FL methods, the global model acts as a teacher that provides soft labels to guide each client's model. This allows clients to absorb global knowledge while retaining personalization, thereby preventing the direct replacement of local model parameters. However, approaches\cite{shenFederatedMutualLearning2023} that rely solely on soft labels from the averaged global model often suffer from anisotropic drift, especially in highly heterogeneous scenarios such as seizure prediction, where EEG distributions vary widely across subjects.

To address these limitations, we propose FedL2T, a novel personalized federated learning framework based on a two-teacher knowledge distillation paradigm (L2T). FedL2T enables each client to learn from both a global teacher and a dynamically assigned peer teacher. Specifically, each client acts not only as a student but also as a potential teacher for others, facilitating direct client-to-client knowledge transfer. This design reduces reliance on a single global teacher and enhances representation diversity. By offering diverse guidance, the peer teacher enables exploration beyond local optima, akin to mutation in genetic algorithms, which helps prevent premature convergence\cite{sastryGeneticAlgorithms2005}. Meanwhile, the global teacher provides stable supervision and helps prevent overfitting. In addition, FedL2T performs hybrid distillation using soft labels and intermediate feature representations. The distillation strength is adaptively adjusted by prediction confidence, emphasizing reliable samples while down-weighting uncertain ones. This collaborative framework supports mutual learning between each client and the global teacher, as well as one-way knowledge transfer from peer clients. To further improve training stability under heterogeneous conditions, a regularization mechanism is introduced to constrain personalized model updates within a controlled range relative to the global model. Finally, we evaluate FedL2T on the public CHB-MIT dataset and a private real-world Renji dataset. Experimental results show that FedL2T consistently outperforms existing methods with faster and more stable convergence under fully supervised settings. Furthermore, it maintains high accuracy across low-label conditions, demonstrating strong robustness to label scarcity and underscoring its practical applicability and potential for real-world clinical deployment.

\begin{figure}[t]  
    \centering 
    \includegraphics[width=0.5\textwidth]{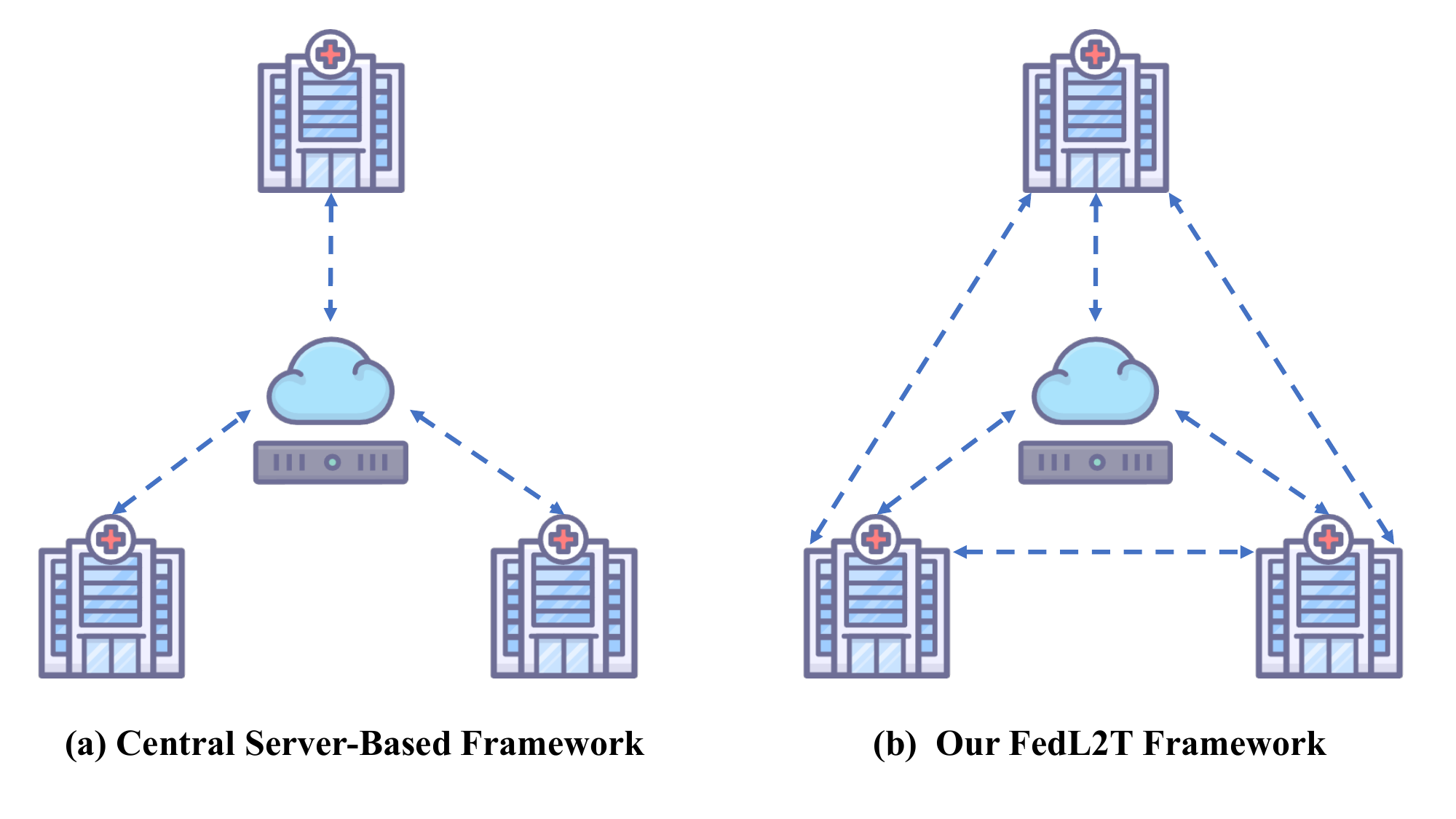} 
    \caption{Comparison of traditional FL and the proposed FedL2T framework. 
    \textbf{(a)} Traditional FL relies on a central server for model aggregation, limiting personalization under heterogeneous data.
    \textbf{(b)} FedL2T introduces an additional peer teacher for distillation, enabling both global and cross-client knowledge transfer to promote learning diversity.}
    \label{fig:Fig1_vs}
\end{figure}

\section{Related Work}
Federated learning has emerged as a promising approach for EEG-based health monitoring, enabling collaborative training of seizure detection and prediction models across distributed hospitals without sharing patient data. Baghersalimi et al.\cite{baghersalimiPersonalizedRealTimeFederated2022} applied the standard FedAvg algorithm to train the ECG-based seizure detection model on edge devices, demonstrating improved accuracy and energy efficiency. Saemaldahr and Ilyas~\cite{saemaldahrPatientSpecificPreictalPatternAware2023a} extended FedAvg within a three-tier architecture, integrating Spiking-GCNN and fuzzy inference to model the patient-specific seizure prediction task. To reduce communication overhead and central dependency, Ding et al.~\cite{FedESDFederatedLearning2023} proposed Fed-ESD for seizure detection, using fog nodes as dynamic aggregators in a hierarchical FL framework. Baghersalimi et al.~\cite{baghersalimiDecentralizedFederatedLearning2024} further extended this direction with a fully decentralized FL approach, combining adaptive model assembly and knowledge distillation to address non-IID challenges. Most recently, Ding et al.~\cite{dingFedMWADModulewiseWeighted2025} presented FedMWAD, which uses module-wise weighted aggregation and personalized local updates to improve generalization while preserving individual adaptability across heterogeneous EEG sources for seizure prediction.

While most existing studies have concentrated on seizure detection, research on federated seizure prediction remains relatively limited. Moreover, the effectiveness of personalized patient-specific modeling still leaves room for improvement. To advance this line of research, we propose FedL2T, which, to our knowledge, is the first framework to introduce a two-teacher knowledge distillation strategy for personalized seizure prediction in a federated setting. By allowing each client to learn jointly from a global model and a dynamically assigned peer model, FedL2T improves model diversity and adaptability without sacrificing privacy or convergence stability. Extensive experiments demonstrate that FedL2T achieves superior classification performance compared to state-of-the-art FL methods.

\begin{figure*}[t]  
    \centering 
    \includegraphics[width=\textwidth]{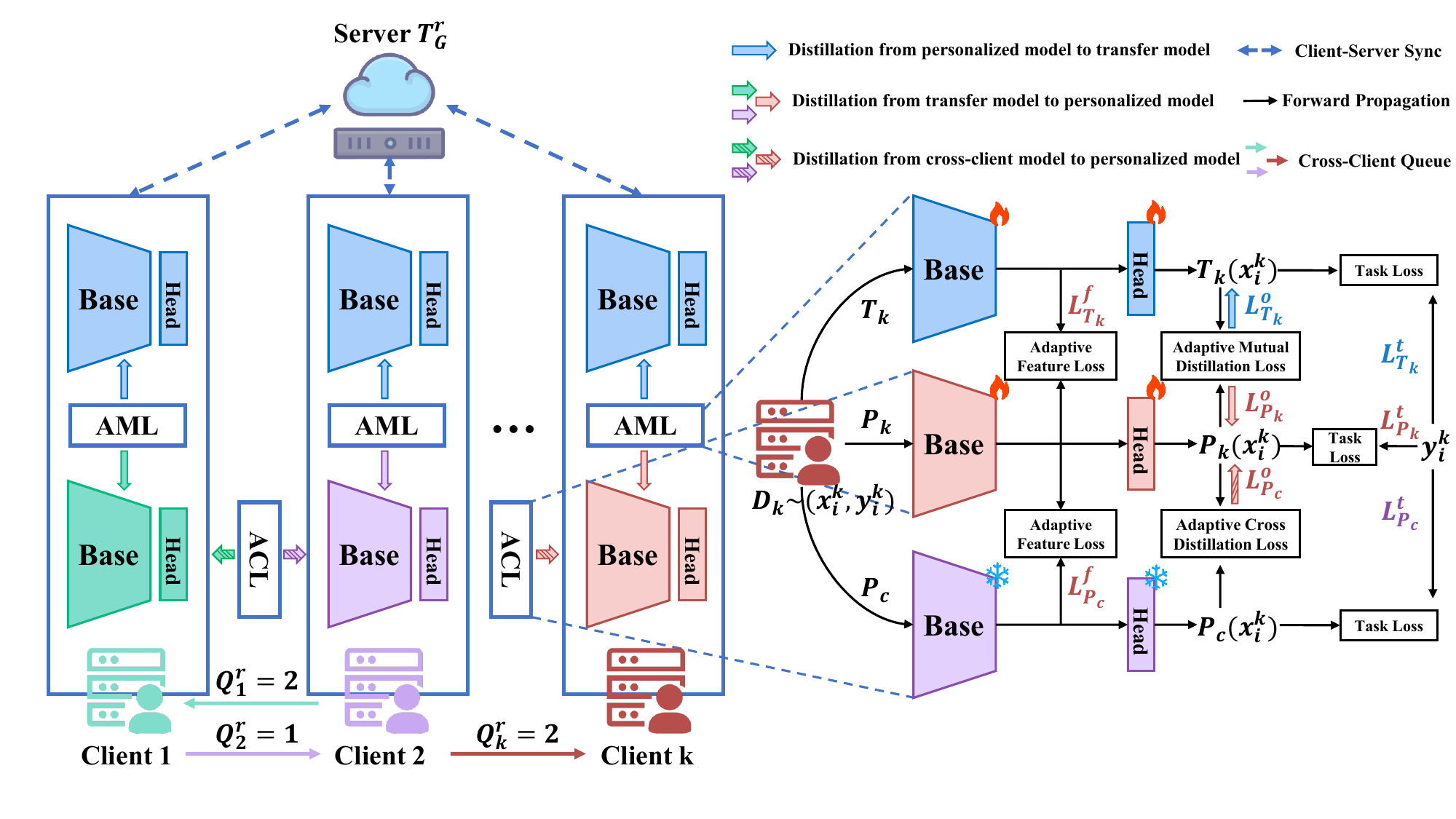} 
    \caption{Overview of the proposed FedL2T framework. Each client maintains a personalized model $\boldsymbol{P}_k$ and a transfer model $\boldsymbol{T}_k$. During local training, FedL2T performs two complementary knowledge distillation processes: Adaptive Mutual Learning (AML) between $\boldsymbol{P}_k$ and $\boldsymbol{T}_k$, and Adaptive Cross Learning (ACL) from a peer model $\boldsymbol{P}_c$ assigned via a cross-client queue $\mathcal{Q}^r$. Both soft predictions and intermediate features are leveraged for multi-level distillation. The transfer models $\boldsymbol{T}_k$ are periodically synchronized with the global model $\boldsymbol{T}^r_G$ aggregated on the server, enabling global coordination without sharing raw data. Colored arrows indicate the direction of knowledge transfer. Flame and snowflake icons respectively indicate updated and frozen parameters, with the latter excluded from gradient updates during training.}
    \label{fig:Fig2_Overview}
\end{figure*}

\section{Methods}
\subsection{Problem Definition}  
Given $K$ clients, each with a private labeled dataset $D_k=\left\{\left(x_i^k, y_i^k\right)\right\}_{i=1}^{n_k}$, which remains locally stored and is never shared. In the FedL2T framework, each client maintains two structurally identical models: a personalized model $\boldsymbol{P}$ tailored to its own data, and a transfer model $\boldsymbol{T}$, periodically synchronized with the global model $\boldsymbol{T}_G^r$ aggregated from all client transfer models. The goal of FedL2T is to collaboratively train robust personalized models across clients while preserving data privacy.

\subsection{Overview of the FedL2T Framework}
Traditional KD-based FL methods typically rely on a single global teacher aggregated by averaging model parameters across clients. While distillation helps prevent the complete overwriting of local knowledge, it often falls short in highly heterogeneous tasks such as personalized EEG-based seizure prediction, where the global model fails to capture patient-specific characteristics. To address this, we propose the FedL2T framework, inspired by the mutation mechanism in genetic algorithms\cite{sastryGeneticAlgorithms2005}, which improves population diversity to escape local optima. FedL2T supplements the global teacher with a dynamically assigned peer teacher from another client during training, promoting richer and more diverse knowledge transfer. An overview of the FedL2T framework is illustrated in Fig.~\ref{fig:Fig2_Overview}.

\subsubsection{Cross-Client Distillation Pathway}
To enable client-to-client knowledge sharing, FedL2T establishes a cross-client distillation pathway in each communication round. The cross-client queue is denoted as $\mathcal{Q}^r=\left\{\mathcal{Q}_1^r, \mathcal{Q}_2^r, \ldots, \mathcal{Q}_k^r\right\}_{k=1}^{K}$, where $\mathcal{Q}_k^r$ represents the peer teacher for client $k$ in round $r$. It is generated randomly each round under the constraint $\mathcal{Q}_k^r \ne k$, allowing each client to learn from a different peer and introducing representational diversity into the training process. An example queue $\left\{2, 1, \ldots, 2\right\}$ in Fig.~2 illustrates a scenario with three clients. Once $\mathcal{Q}^r$ is determined, all clients update in parallel. The update includes two components: Adaptive Cross Learning (ACL) and Adaptive Mutual Learning (AML), which are described in the following subsections.

\subsubsection{Adaptive Cross Learning}
Given $\mathcal{Q}^r_k = c$, client $k$ receives the $\boldsymbol{P}_c$ from client $c$ and performs one-way knowledge distillation from $\boldsymbol{P}_c$ to its personalized model $\boldsymbol{P}_k$. Acting as a cross-client teacher, $\boldsymbol{P}_c$ provides external knowledge through multi-level distillation, thereby enhancing the generalization capability of $\boldsymbol{P}_k$. In addition to distillation, $\boldsymbol{P}_k$ is optimized on its private dataset $D_k$ using cross-entropy loss to fit the ground truth labels. This local supervision ensures that $\boldsymbol{P}_k$ retains its base performance and mitigates the risk of client drift. Meanwhile, $\boldsymbol{P}_c$ performs forward inference on $D_k$ with frozen parameters and remains unchanged. For each training sample $(x_i^k, y_i^k) \in D_k$, let $\boldsymbol{P}_k(x_i^k)$ and $\boldsymbol{P}_c(x_i^k)$ denote the soft probabilities predicted by the personalized and cross-client models, respectively. The task losses $\mathcal{L}_{{P}_k}^t$ and $\mathcal{L}_{{P}_c}^t$ are computed as:
\begin{equation}
\mathcal{L}_{{P}_k}^t = \mathcal{L}_{CE}\left(\boldsymbol{P}_k(x_i^k), y_i^k\right), \;
\mathcal{L}_{{P}_c}^t = \mathcal{L}_{CE}\left(\boldsymbol{P}_c(x_i^k), y_i^k\right)
\label{eq:CL}
\end{equation}
where $\mathcal{L}_{CE}(p, y) = -[y \log p + (1 - y) \log (1 - p)]$ denotes the binary cross-entropy loss, which is widely used in seizure prediction to quantify the discrepancy between predicted probabilities and binary ground-truth labels.

Previous studies \cite{romeroFitNetsHintsThin2015, 2020MiniLM} have shown that incorporation of knowledge distillation at the feature level can provide a finer-grained and more comprehensive transfer of information between models. Thus, we adopt a multi-level distillation strategy comprising both output-level and feature-level alignment to enhance cross-client knowledge transfer. Specifically, the output-level loss $\mathcal{L}_{{P}c}^{o}$ corresponds to the Adaptive Cross Distillation Loss module in Fig.~\ref{fig:Fig2_Overview}, while the feature-level loss $\mathcal{L}_{{P}_c}^{f}$ corresponds to the Adaptive Feature Loss module. These losses are defined as follows:
\begin{equation}
\mathcal{L}_{{P}_c}^{o} = \mathcal{L}_{KL}\left(\boldsymbol{P}_c(x_i^k) \, \| \, \boldsymbol{P}_k(x_i^k)\right) \ / \ ({\mathcal{L}_{{P}_c}^t + \mathcal{L}_{{P}_k}^t})
\label{eq:KL}
\end{equation} 
\begin{equation}
\mathcal{L}_{{P}_c}^{f} = \mathcal{L}_{MSE}\left(\boldsymbol{P}_{cb}(x_i^k), \boldsymbol{P}_{kb}(x_i^k)\right)\ / \  ({\mathcal{L}_{{P}_c}^t + \mathcal{L}_{{P}_k}^t})
\label{eq:MSE}
\end{equation} 
where $\mathcal{L}_{KL}$ denotes the Kullback–Leibler divergence between the predicted soft probabilities of the teacher $\boldsymbol{P}_c$ and the student $\boldsymbol{P}_k$, and $\mathcal{L}_{MSE}$ computes the mean squared error between their intermediate feature representations extracted by the base block ($b$) of model $\boldsymbol{P}$ (detailed in Section~\ref{sec:Model Architecture}). Furthermore, to achieve adaptive regulation of knowledge transfer, we adopt the strategy proposed in FedKD\cite{wuCommunicationefficientFederatedLearning2022}, where both distillation losses are normalized by the sum of task losses. This design ensures that, when task losses are large, indicating less reliable predictions, the impact of distillation is reduced to prevent the propagation of low-confidence knowledge. In contrast, when both models achieve low task losses, the normalized distillation loss increases accordingly, which enhances the transfer of reliable knowledge and helps mitigate overfitting.

\subsubsection{Adaptive Mutual Learning}
Complementary to ACL, which introduces diversity via cross-client knowledge transfer, AML facilitates mutual knowledge exchange\cite{zhangDeepMutualLearning2018a, shenFederatedMutualLearning2023} between the local transfer model $\boldsymbol{T}_k$ and the personalized model $\boldsymbol{P}_k$. This bidirectional interaction enables $\boldsymbol{P}_k$ to absorb global knowledge from $\boldsymbol{T}_k$, while $\boldsymbol{T}_k$, enriched with local task-specific information, is periodically uploaded to the server and aggregated into the global model $\boldsymbol{T}^r_G$. Specifically, during each local epoch, both $\boldsymbol{P}_k$ and $\boldsymbol{T}_k$ are updated using local data $D_k$ while engaging in an adaptive mutual distillation process to exchange knowledge bidirectionally. The output-level distillation loss from $\boldsymbol{P}_k$ to $\boldsymbol{T}_k$, corresponding to the Adaptive Mutual Distillation Loss module in Fig.~\ref{fig:Fig2_Overview}, is defined as:
\begin{equation}
\mathcal{L}_{{T}_k}^{o} = \mathcal{L}_{KL}\left(\boldsymbol{P}_k(x_i^k) \, \| \, \boldsymbol{T}_k(x_i^k)\right) / \left(\mathcal{L}_{{P}_k}^t + \mathcal{L}_{{T}_k}^t\right)
\end{equation}
where $\mathcal{L}_{{P}_k}^t$ and $\mathcal{L}_{{T}_k}^t$ represent the task losses of $\boldsymbol{P}_k$ and $\boldsymbol{T}_k$, respectively, both computed using the formulation in Eq.~\eqref{eq:CL} with the corresponding model substituted. The remaining losses, including the output-level distillation loss $\mathcal{L}_{{P}_k}^{o}$ (from $\boldsymbol{T}_k$ to $\boldsymbol{P}_k$) and the feature-level loss $\mathcal{L}_{{T}_k}^{f}$, are computed following Eqs.~\eqref{eq:KL} and \eqref{eq:MSE}, with $\boldsymbol{T}_k$ substituted for $\boldsymbol{P}_c$. 

Although the two-teacher strategy comprising ACL and AML facilitates both diversity and global knowledge integration, data heterogeneity may cause personalized models to deviate considerably from the global model. This deviation often leads to severe fluctuations during distillation, resulting in unstable training and poor generalization. To mitigate this issue, we introduce a regularization mechanism inspired by \cite{liDittoFairRobust2021}, which penalizes the divergence of $\boldsymbol{P}_k$ from the global model $\boldsymbol{T}^r_G$. The proximal term is formulated as:
\begin{equation}
\mathcal{L}_{\text{prox}} = \frac{\mu}{2} \left\| \boldsymbol{P}_k - \boldsymbol{T}^r_G \right\|_2^2
\end{equation}
where $\mu$ is a tunable hyperparameter that controls the strength of regularization. 

\subsubsection{Client Update and Server Aggregation}
Finally, the total loss of client $k$ in FedL2T at each epoch is formulated as:
\begin{equation}
\mathcal{L}_{T_k} = \mathcal{L}^{t}_{T_k} + (\mathcal{L}^{o}_{T_k} + \mathcal{L}^{f}_{T_k})
\end{equation}
\begin{equation}
\mathcal{L}_{P_k} = \mathcal{L}^{t}_{P_k} + (\mathcal{L}^{o}_{P_k} + \mathcal{L}^{f}_{T_k}) + \lambda_{c}(\mathcal{L}^{o}_{P_c} + \mathcal{L}^{f}_{P_c}) + 
\mathcal{L}_{prox}
\end{equation}
where $\boldsymbol{T}_k$ is optimized using task supervision and distillation from $\boldsymbol{P}_k$, while $\boldsymbol{P}_k$ incorporates an additional proximal term and cross-client distillation from $\boldsymbol{P}_c$, weighted by $\lambda_c$ to balance contributions from both teachers. Subsequently, $\boldsymbol{T}_k$ and $\boldsymbol{P}_k$ are updated by gradient descent:
\begin{align}
\boldsymbol{T}_k = \boldsymbol{T}_k - \eta \nabla_{{T}_k} \mathcal{L}_{T_k}, \ 
\boldsymbol{P}_k = \boldsymbol{P}_k - \eta \nabla_{{P}_k} \mathcal{L}_{P_k}
\end{align}
where $\eta$ denotes the learning rate. The gradients $\nabla_{{T}_k} \mathcal{L}_{T_k}$ and $\nabla_{{P}_k} \mathcal{L}_{P_k}$ are computed with respect to the total loss of $\boldsymbol{T}_k$ and $\boldsymbol{P}_k$, respectively.

After $E$ epochs of local updates, each client uploads its optimized $\boldsymbol{T}_k$ to the server, which aggregates all transfer models into a global model $\boldsymbol{T}^{r+1}_G$ by weighted averaging:
\begin{equation}
\boldsymbol{T}^{r+1}_{G} = \sum_{k=1}^{K} \frac{n_k}{n} \boldsymbol{T}_k, \quad \text{where} \; n = \sum_{k=1}^{K} n_k
\end{equation}

This aggregated model is then broadcast as the new initialization for $\boldsymbol{T}_k$ in the next round. Through global updates of $\boldsymbol{T}_k$ and local training of $\boldsymbol{P}_k$, FedL2T enables personalized optimization while preserving data privacy. The detailed steps of the proposed FedL2T algorithm are presented in Algorithm~\ref{alg: alg1}.

\begin{algorithm}[t]
    \caption{The FedL2T Algorithm}
    \label{alg: alg1}
    \textbf{Input:} Total communication rounds $R$, Local epochs $E$ \\
    \hspace*{9.8mm} Total number of clients $K$ \\
    \hspace*{9.8mm} The network structure for $\boldsymbol{P}$, $\boldsymbol{T}$ \\
    \textbf{Output:} The trained personalized local models $\left\{ \boldsymbol{P}_k \right\}_{k=1}^{K}$
    \textbf{Server executes:}
    \begin{algorithmic}[1]
    \State Initialize the model parameters $\boldsymbol{P}$, $\boldsymbol{T}$ for all clients
    \For{each round $r$ in $R$}
        \For{each client $k$ \textbf{in parallel}}
            \State Assign random cross-client teacher $\boldsymbol{P}_c$
            \State $\boldsymbol{T}_k \leftarrow$
            \textbf{ClientUpdate}$(k, \boldsymbol{P}_k, \boldsymbol{P}_c)$
        \EndFor 
        \State Update global model:  
        $\boldsymbol{T}^{r+1}_{G} \leftarrow \sum_{k=1}^{K} \frac{n_k}{n} \boldsymbol{T}_k$

        \State Broadcast $\boldsymbol{T}^{r+1}_{G}$ to all $\boldsymbol{T}_k$
        
    \EndFor
    \end{algorithmic}
    \textbf{ClientUpdate$(k, \boldsymbol{P}_k, \boldsymbol{P}_c)$:}
    \begin{algorithmic}[1]
    \For{each local epoch $e$ in $E$}
        \State Compute task losses $\mathcal{L}^{t}_{P_k}$, $\mathcal{L}^{t}_{T_k}$ and $\mathcal{L}^{t}_{P_c}$
        \State Compute output distillation losses $\mathcal{L}^{o}_{P_k}$, $\mathcal{L}^{o}_{T_k}$ and $\mathcal{L}^{o}_{P_c}$
        \State Compute feature distillation losses $\mathcal{L}^{f}_{T_k}$ and $\mathcal{L}^{f}_{P_c}$
        \State Compute proximal term $\mathcal{L}_{prox}$

        \State $\mathcal{L}_{T_k} \leftarrow \mathcal{L}^{t}_{T_k}$ + $(\mathcal{L}^{o}_{T_k}$ + $\mathcal{L}^{f}_{T_k})$
        \State $\mathcal{L}_{P_k} \leftarrow \mathcal{L}^{t}_{P_k} + (\mathcal{L}^{o}_{P_k} + \mathcal{L}^{f}_{T_k}) + \lambda_{c}(\mathcal{L}^{o}_{P_c} + \mathcal{L}^{f}_{P_c}) + \mathcal{L}_{prox}$

        \State $\boldsymbol{T}_k \leftarrow \boldsymbol{T}_k - \eta \nabla_{{T}_k} \mathcal{L}_{T_k}$
        \State $\boldsymbol{P}_k \leftarrow \boldsymbol{P}_k - \eta \nabla_{{P}_k} \mathcal{L}_{P_k}$
        
    \EndFor
    \State \Return $\boldsymbol{T}_k$
    \end{algorithmic}
\end{algorithm}

\section{Experiments}
This section presents the experimental setup to evaluate our proposed FedL2T framework for epileptic seizure prediction.

\subsection{EEG Datasets}
To evaluate the proposed method, experiments are conducted on both a public EEG dataset and a private EEG dataset, with all recordings following the international 10–20 electrode placement system. The public CHB-MIT dataset\cite{Shoeb2009ApplicationOM} includes long-term scalp EEG recordings from 23 pediatric subjects sampled at 256 Hz, with 21 commonly available electrodes selected as input. The private Renji dataset consists of EEG recordings from 18 adult patients acquired at 200 Hz, using 16 channel configurations. All raw EEG data are preprocessed with notch filtering to suppress power-line noise, followed by z-score normalization. For the CHB-MIT dataset, the continuous signals are segmented into 7-second windows, while the Renji data is segmented using the same number of sampling points with 50\% overlap to augment data samples. Each segment is denoted as $x_i \in \mathbb{R}^{{chs} \times T}$, where ${chs}$ is the number of EEG channels and $T$ is the number of sampling points. In the FL setting, each subject is treated as an individual client with a private dataset $D_k = {(x_i^k, y_i^k)}_{i=1}^{n_k}$.

\begin{table}[t]
    \centering
    \caption{Detailed Description of Subjects Included in the CHB-MIT and Renji Datasets.}
    \label{tab: DATASET}
    \renewcommand{\arraystretch}{1.05}
    \setlength{\tabcolsep}{0.8pt}  
    \resizebox{0.5\textwidth}{!}{ 
        \begin{tabular}{cccccc|cccccc}
            \hline
            \multicolumn{6}{c|}{CHB-MIT}                                              & \multicolumn{6}{c}{Renji}                                                \\ \hline
            Subject & Age  & \multicolumn{1}{c|}{Seizures} & Subject & Age & Seizures & Subject & Age & \multicolumn{1}{c|}{Seizures} & Subject & Age & Seizures \\ \hline
            1       & 11   & \multicolumn{1}{c|}{6}        & 14      & 9   & 4        & S01    & 24  & \multicolumn{1}{c|}{4}        & S10   & 32  & 5        \\
            2       & 11   & \multicolumn{1}{c|}{3}        & 16      & 7   & 5        & S02    & 21  & \multicolumn{1}{c|}{4}        & S11   & 48  & 4        \\
            3       & 14   & \multicolumn{1}{c|}{6}        & 17      & 12  & 3        & S03    & 85  & \multicolumn{1}{c|}{3}        & S12   & 60  & 4        \\
            5       & 7    & \multicolumn{1}{c|}{5}        & 18      & 18  & 6        & S04    & 46  & \multicolumn{1}{c|}{4}        & S13   & 41  & 4        \\
            6       & 1.5  & \multicolumn{1}{c|}{7}        & 19      & 19  & 3        & S05    & 48  & \multicolumn{1}{c|}{3}        & S14   & 21  & 4        \\
            7       & 14.5 & \multicolumn{1}{c|}{3}        & 20      & 6   & 5        & S06    & 17  & \multicolumn{1}{c|}{5}        & S15   & 49  & 3        \\
            8       & 3.5  & \multicolumn{1}{c|}{4}        & 21      & 13  & 4        & S07    & 35  & \multicolumn{1}{c|}{4}        & S16   & 87  & 3        \\
            9       & 10   & \multicolumn{1}{c|}{4}        & 22      & 9   & 3        & S08    & 26  & \multicolumn{1}{c|}{4}        & S17   & 39  & 3        \\
            10      & 3    & \multicolumn{1}{c|}{6}        & 23      & 6   & 4        & S09    & 59  & \multicolumn{1}{c|}{3}        & S18   & 25  & 4        \\ 
            \hline
        \end{tabular}
    }
\end{table}

\begin{table}[t!]
    \centering
    \caption{Details of Model Architecture.}
    \label{tab:cnn_gru_structure}
    \renewcommand{\arraystretch}{1.0}
    \setlength{\tabcolsep}{2pt}  
    \resizebox{0.5\textwidth}{!}{ 
        \begin{tabular}{cccccccc}
        \toprule
        \textbf{Layer} & \textbf{Input} & \textbf{Type} & \textbf{Kernel} & \textbf{Stride} & \textbf{Pool} & \textbf{Stride} & \textbf{Output} \\
        \midrule
        Conv1 & $[chs,\ T]$           & Conv1d & 5   & 2 & 2 & 2 & $[64,\ T//4]$ \\
        Conv2 & $[64,\ T//4]$         & Conv1d & 3   & 2 & 2 & 2 & $[64,\ T//16]$ \\
        Conv3 & $[64,\ T//16]$        & Conv1d & 3   & 1 & 2 & 2 & $[64,\ T//32]$ \\
        Conv4 & $[64,\ T//32]$        & Conv1d & 2   & 1 & 2 & 2 & $[128,\ T//64]$ \\
        GRU   & $[128,\ T//64]^\top$  & GRU    & --  & --& --& --& $[128]$ \\
        FC    & $[128]$               & Linear & --  & --& --& --& $[2]$ \\
        \bottomrule
        \end{tabular}
    }
    \begin{flushleft}
    \footnotesize   
    Note: $//$ denotes integer division; $^\top$ indicates transposition. The GRU contains 128 hidden units, with only the final hidden state used as output.
    \end{flushleft}
\end{table}

The Seizure Occurrence Period (SOP) and Seizure Prediction Horizon (SPH) are set to 30 and 5 minutes, respectively, balancing timely intervention with reduced patient anxiety. Inter-ictal segments are defined as EEG windows occurring at least 4 hours from any seizure to ensure clear separation from ictal states. To address class imbalance, inter-ictal samples are undersampled. To reflect realistic clinical settings where only past data can inform future predictions, a leave-last-event-out strategy is adopted: for each patient with $N$ seizures, the last is used for testing, and the remaining $(N-1)$ seizures are used for training. The detailed subject information for both the CHB-MIT and Renji datasets is summarized in Table~\ref{tab: DATASET}.

\subsection{Evaluation Metrics}
As the experimental results report subject-wise performance, segment-based accuracy (Acc) is adopted as the primary evaluation metric for a concise comparison. Accuracy, defined as the proportion of correctly classified segments among all segments, offers a reliable measure of model performance under class-balanced conditions.

\subsection{Model Architecture}
\label{sec:Model Architecture}
The model architecture employed in our experiments is summarized in Table~\ref{tab:cnn_gru_structure}. Each model comprises two components: a base block for feature extraction and a head block for classification. The base block consists of four consecutive convolutional modules followed by a Gated Recurrent Unit (GRU) layer. Each convolutional module includes a 1D convolutional layer, batch normalization, a LeakyReLU activation, and a max-pooling layer, which together progressively reduce the temporal resolution. The GRU layer models temporal dependencies within the compressed sequences. The head block is a fully connected (FC) layer that projects the extracted features into a binary prediction space for seizure classification.

\subsection{Implementation Details}
All experiments are implemented in PyTorch and run on a system with an NVIDIA GeForce RTX 4090 GPU and Ubuntu OS. The models are optimized using stochastic gradient descent (SGD) with a learning rate of $\eta = 0.01$. The regularization coefficient $\mu$ is set to $0.2$, and the cross-client distillation weight $\lambda_c$ is set to $0.5$. A total of $R = 100$ communication rounds are performed, with each client conducting $E = 1$ local epoch per round.

\begin{table*}[t!]
    \centering
    \caption{Comparison with State-of-the-art and Ablation Variants on CHB-MIT Dataset.} 
    \renewcommand{\arraystretch}{1.05}
    \label{tab: Fed-CHBMIT}
    \resizebox{1.0\textwidth}{!}{
        \begin{tabular}{c|ccccccc|ccc}
            \hline
            Subject                         & Local & FedAvg\cite{McMahanMRHA17} & FedBN\cite{liFedBNFederatedLearning2020} & FedRep\cite{collinsExploitingSharedRepresentations2021} & FedMRL\cite{yi2024FedMRL} & Ditto\cite{liDittoFairRobust2021}  & FML\cite{shenFederatedMutualLearning2023}    & L2T-C & L2T-CG & FedL2T (Ours) \\ \hline
            1                          & 96.89  & 88.91  & 98.44 & 97.28  & 96.50  & 96.89  & 94.94  & 95.91  & 97.47  & 100.00 \\
            2                          & 70.04  & 57.98  & 99.61 & 98.05  & 70.43  & 96.89  & 83.66  & 82.88  & 66.54  & 99.61  \\
            3                          & 95.91  & 85.41  & 84.63 & 94.75  & 98.83  & 92.80  & 97.28  & 98.64  & 99.61  & 97.67  \\
            5                          & 99.03  & 95.33  & 94.75 & 98.25  & 99.22  & 99.81  & 97.08  & 99.42  & 100.00 & 99.81  \\
            6                          & 77.24  & 79.77  & 95.33 & 64.79  & 66.54  & 99.81  & 58.95  & 65.56  & 100.00 & 99.81  \\
            7                          & 93.39  & 38.13  & 96.11 & 99.81  & 80.16  & 90.47  & 94.94  & 98.25  & 93.00  & 98.83  \\
            8                          & 96.69  & 89.69  & 99.42 & 100.00 & 98.44  & 99.42  & 97.86  & 99.22  & 100.00 & 100.00 \\
            9                          & 100.00 & 70.04  & 99.61 & 96.11  & 99.81  & 100.00 & 99.03  & 100.00 & 100.00 & 100.00 \\
            10                         & 96.50  & 55.45  & 91.25 & 99.61  & 100.00 & 99.81  & 99.61  & 100.00 & 100.00 & 99.81  \\
            14                         & 70.04  & 56.23  & 97.67 & 92.22  & 92.02  & 100.00 & 91.25  & 98.64  & 100.00 & 100.00 \\
            16                         & 92.41  & 93.77  & 98.64 & 93.19  & 98.83  & 96.69  & 93.97  & 98.25  & 100.00 & 94.36  \\
            17                         & 93.97  & 49.22  & 69.84 & 97.08  & 89.69  & 58.17  & 92.02  & 92.80  & 97.47  & 81.13  \\
            18                         & 99.42  & 71.79  & 81.52 & 99.42  & 99.61  & 99.22  & 98.05  & 99.81  & 100.00 & 100.00 \\
            19                         & 99.22  & 50.00  & 89.11 & 100.00 & 99.03  & 93.39  & 99.03  & 99.42  & 99.81  & 100.00 \\
            20                         & 100.00 & 59.14  & 96.89 & 99.81  & 97.67  & 96.30  & 100.00 & 99.81  & 100.00 & 100.00 \\
            21                         & 94.36  & 55.06  & 50.00 & 54.28  & 93.00  & 90.47  & 87.94  & 84.44  & 98.05  & 93.00  \\
            22                         & 88.72  & 84.63  & 99.42 & 97.67  & 93.77  & 99.81  & 99.42  & 98.83  & 97.08  & 100.00 \\
            23                         & 100.00 & 63.04  & 99.81 & 100.00 & 99.81  & 100.00 & 100.00 & 100.00 & 100.00 & 100.00 \\ \hline
            {Avg} & 92.44 & 69.09\textcolor{red}{↓23.35} & 91.23\textcolor{red}{↓1.21} & 93.46\textcolor[rgb]{0,0.5,0}{↑1.02}  & 92.96\textcolor[rgb]{0,0.5,0}{↑0.52} & 95.00\textcolor[rgb]{0,0.5,0}{↑2.56} & 93.61\textcolor[rgb]{0,0.5,0}{↑1.17} & 95.10\textcolor[rgb]{0,0.5,0}{↑2.66} & 97.17\textcolor[rgb]{0,0.5,0}{↑4.73} & \textbf{98.00}\textcolor[rgb]{0,0.5,0}{↑5.56} \\ \hline
        \end{tabular}
    }
    
    \begin{flushleft}
        \footnotesize        
        Note: All values represent classification accuracy (\%) for each subject. The "Avg" row reports the average accuracy across all subjects, where \textbf{Bold} denotes the highest averaged accuracy. The up\textcolor[rgb]{0,0.5,0}{↑} and down\textcolor{red}{↓} arrows represent the improvement or decline relative to the Local baseline. 
    \end{flushleft}
\end{table*}

\begin{table*}[t]
    \centering
    \caption{Comparison with State-of-the-art and Ablation Variants on Renji Dataset.} 
    \renewcommand{\arraystretch}{1.05}
    \label{tab: Fed-Private}
    \resizebox{1.0\textwidth}{!}{
        \begin{tabular}{c|ccccccc|ccc}
            \hline
            Subject                         & Local  & FedAvg\cite{McMahanMRHA17} & FedBN\cite{liFedBNFederatedLearning2020}  & FedRep\cite{collinsExploitingSharedRepresentations2021} & FedMRL\cite{yi2024FedMRL} & Ditto\cite{liDittoFairRobust2021} & FML\cite{shenFederatedMutualLearning2023}    & L2T-C & L2T-CG & FedL2T (Ours)\\ \hline
            S01                         & 92.98  & 67.17 & 85.34  & 73.81  & 88.85  & 93.36 & 97.99  & 96.62 & 97.74  & 93.61  \\
            S02                         & 99.37  & 75.69 & 100.00 & 100.00 & 99.75  & 92.48 & 99.00  & 97.24 & 99.00  & 99.50  \\
            S03                         & 96.74  & 95.99 & 99.75  & 100.00 & 100.00 & 99.37 & 97.49  & 97.37 & 99.12  & 99.50  \\
            S04                         & 51.88  & 67.29 & 55.64  & 50.13  & 83.46  & 66.79 & 88.60  & 84.46 & 85.09  & 82.32  \\
            S05                         & 97.87  & 69.80 & 99.50  & 99.75  & 100.00 & 97.62 & 99.37  & 99.25 & 99.12  & 98.87  \\
            S06                         & 98.50  & 86.59 & 99.75  & 100.00 & 99.87  & 78.95 & 100.00 & 99.75 & 99.75  & 99.87  \\
            S07                         & 91.48  & 57.64 & 86.47  & 91.73  & 93.23  & 79.32 & 68.42  & 83.83 & 87.72  & 91.85  \\
            S08                         & 75.94  & 46.62 & 65.16  & 56.89  & 84.71  & 81.33 & 93.98  & 94.74 & 91.85  & 96.92  \\
            S09                         & 92.98  & 88.22 & 93.73  & 95.74  & 94.74  & 94.36 & 96.24  & 97.62 & 97.99  & 98.25  \\
            S10                        & 92.11  & 76.32 & 96.49  & 77.32  & 86.22  & 98.62 & 50.50  & 93.48 & 94.99  & 93.11  \\
            S11                        & 97.49  & 91.23 & 99.25  & 92.61  & 95.74  & 99.75 & 95.11  & 97.49 & 98.50  & 99.25  \\
            S12                        & 87.72  & 81.08 & 97.87  & 99.87  & 97.74  & 89.10 & 98.25  & 93.48 & 94.61  & 99.50  \\
            S13                        & 91.48  & 92.48 & 92.23  & 99.50  & 92.36  & 95.36 & 95.99  & 99.00 & 99.25  & 97.12  \\
            S14                        & 84.71  & 70.80 & 94.61  & 93.23  & 91.48  & 82.46 & 92.98  & 98.12 & 95.49  & 88.10  \\
            S15                        & 96.49  & 72.56 & 70.68  & 87.72  & 93.11  & 98.25 & 97.99  & 96.12 & 99.62  & 99.12  \\
            S16                        & 96.74  & 81.70 & 88.97  & 98.87  & 97.87  & 91.35 & 100.00 & 98.12 & 98.75  & 99.00  \\
            S17                        & 88.47  & 96.74 & 99.37  & 70.93  & 79.57  & 99.62 & 99.62  & 68.92 & 86.97  & 99.50  \\
            S18                        & 99.25  & 99.87 & 99.75  & 89.85  & 89.97  & 99.87 & 90.10  & 99.75 & 99.75  & 99.75  \\ \hline
            {Avg} & 90.68 & 78.77\textcolor{red}{↓11.91} & 90.25\textcolor{red}{↓0.43} & 87.66\textcolor{red}{↓3.02} & 92.70\textcolor[rgb]{0,0.5,0}{↑2.02} & 91.00\textcolor[rgb]{0,0.5,0}{↑0.32} & 92.31\textcolor[rgb]{0,0.5,0}{↑1.63} & 94.19\textcolor[rgb]{0,0.5,0}{↑3.51} & 95.85\textcolor[rgb]{0,0.5,0}{↑5.17} & \textbf{96.40}\textcolor[rgb]{0,0.5,0}{↑5.72} \\ \hline
        \end{tabular}
    }

    \begin{flushleft}
        \footnotesize
        Note: All values represent classification accuracy (\%) for each subject. The "Avg" row reports the average accuracy across all subjects, where \textbf{Bold} denotes the highest averaged accuracy. The up\textcolor[rgb]{0,0.5,0}{↑} and down\textcolor{red}{↓} arrows represent the improvement or decline relative to the Local baseline. 
    \end{flushleft}
\end{table*}

\label{sec:experiments}

\section{Results and Discussion}

\subsection{Comparison with State-of-the-art Methods}
To evaluate the effectiveness of the proposed FedL2T framework, we conduct extensive experiments against several representative FL methods, including FedAvg\cite{McMahanMRHA17}, FedBN\cite{liFedBNFederatedLearning2020}, FedRep\cite{collinsExploitingSharedRepresentations2021}, FedMRL\cite{yi2024FedMRL}, Ditto\cite{liDittoFairRobust2021}, and FML\cite{shenFederatedMutualLearning2023}, as well as the Local baseline where each client is trained independently without collaboration, i.e., patient-specific training strategy commonly adopted in the seizure prediction task. 

As shown in Table~\ref{tab: Fed-CHBMIT}, our proposed FedL2T achieves the highest average accuracy of 98.00\% on the CHB-MIT dataset, significantly outperforming all competing methods. Notably, it surpasses the Local baseline by 5.56\%, demonstrating its ability to retain personalization while benefiting from cross-client knowledge transfer. Similar advantages are observed on the Renji dataset in Table~\ref{tab: Fed-Private}, where FedL2T reaches 96.40\% average accuracy, exceeding the Local baseline by 5.72\% and consistently outperforming all other methods. These results validate the robustness and generalizability of the proposed two-teacher learning paradigm strategy across heterogeneous datasets.

Serving as a lower bound for FL performance, FedAvg achieves average accuracies of 69.09\% and 78.77\% on CHB-MIT and the Renji datasets, respectively. In contrast, all five personalized FL methods exhibit notable improvements by incorporating different personalization strategies. FedBN applies local batch normalization to mitigate feature shift, FedRep decouples shared representation learning from personalized classification heads, and FedMRL introduces a nested model design to accommodate structural heterogeneity. Moreover, Ditto incorporates a regularization-based dual model strategy to stabilize local adaptation, while FML employs mutual knowledge distillation to transfer global knowledge. However, their performance relative to the Local baseline is inconsistent. While FedMRL, Ditto and FML generally achieve slight improvements over Local on both datasets, FedBN and FedRep occasionally underperform. This inconsistency highlights the challenge of stable personalization and further emphasizes the effectiveness of FedL2T in handling data heterogeneity.

\begin{figure*}[t!]
    \centering      
    \includegraphics[width=1.0\textwidth]{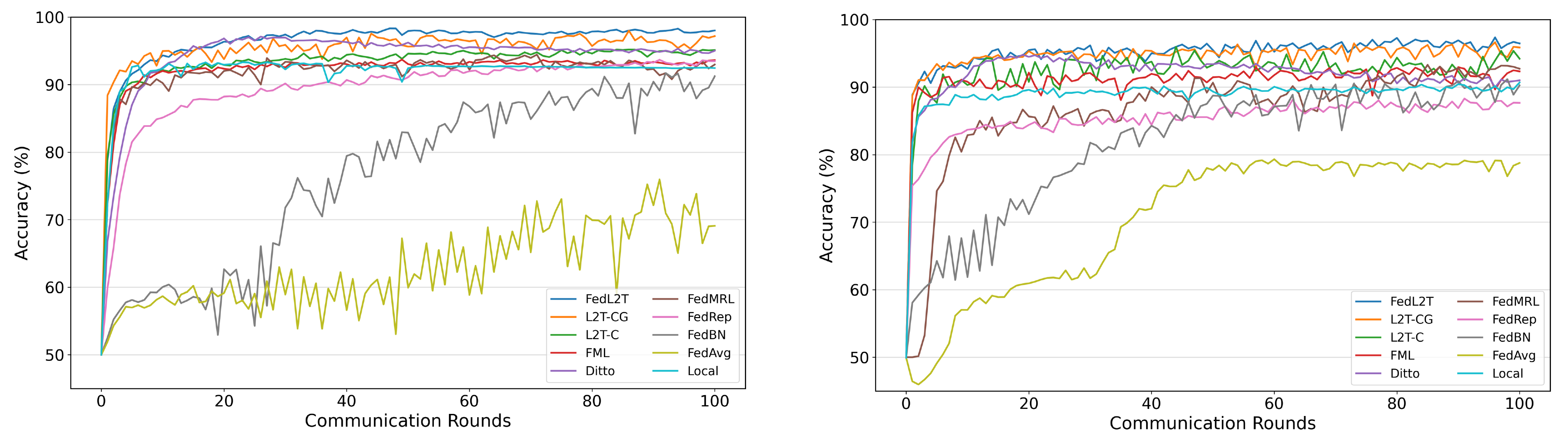}  
    \caption{Comparison of classification accuracy (\%) across communication rounds on the CHB-MIT (left) and Renji (right) datasets. Each curve represents the average test accuracy of a federated method across clients.}
    \label{fig:Fig3_curve}
\end{figure*}

\begin{table*}[t]
    \caption{Accuracy (\%) Comparison under Varying Labeled Data Ratios (\%) on Two Datasets. “Mean ± Std” Represents the Average Accuracy and Standard Deviation Across 18 Subjects.} 
    \centering
    \renewcommand{\arraystretch}{1.1}
    \label{tab: Semi_Labels}
    \resizebox{1.0\textwidth}{!}{
        \begin{tabular}{c|cccc|cccc}
        \hline
                                  & \multicolumn{4}{c|}{CHB-MIT}                                                                                                         & \multicolumn{4}{c}{{Renji}}                                                                   \\ \cline{2-9} 
        \multirow{-2}{*}{Methods} & {1\%   Labels} & {5\% Labels} & {10\% Labels} & 25\% Labels           & {1\%   Labels}          & 5\% Labels            & 10\% Labels           & 25\% Labels           \\ \hline
        Local                     & 73.99 ± 17.46                       & 86.10 ± 13.31                     & 89.66 ± 13.95                      & 90.43 ± 13.57         & {81.60 ± 12.83}         & 85.90 ± 10.60         & 87.52 ± 10.57         & 88.62 ± 10.88         \\
        FedAvg\cite{McMahanMRHA17}                    & 52.84 ± 12.42                       & 54.17 ± 8.36                      & 63.80 ± 10.09                      & 66.75 ± 22.95         & {49.65 ± 2.54}          & 50.49 ± 15.12         & 55.53 ± 10.26         & 61.48 ± 23.52         \\
        FedBN\cite{liFedBNFederatedLearning2020}                     & 58.84 ± 13.76                       & 71.66 ± 10.43                     & 82.77 ± 12.36                      & 88.63 ± 13.31         & {79.35 ± 12.18}         & 84.45 ± 12.21         & 87.01 ± 11.78         & 89.55 ± 12.85         \\
        FedRep\cite{collinsExploitingSharedRepresentations2021}                    & 69.53 ± 15.16                       & 79.83 ± 16.47                     & 86.42 ± 14.05                      & 90.24 ± 13.08         & {81.84 ± 14.99}         & 83.88 ± 14.85         & 85.55 ± 15.15         & 86.32 ± 15.68         \\
        FedMRL\cite{yi2024FedMRL}                    & 72.56 ± 9.05                        & 82.34 ± 15.56                     & 90.30 ± 12.10                      & 91.06 ± 11.63         & {82.05 ± 15.45}         & 85.89 ± 14.00         & 88.43 ± 13.32         & 91.73 ± 8.00          \\
        Ditto\cite{liDittoFairRobust2021}                     & 80.56 ± 12.70                       & 86.15 ± 11.25                     & 89.20 ± 10.13                      & 91.49 ± 9.16          & {83.10 ± 11.33}         & 86.58 ± 11.45         & 87.80 ± 11.75         & 90.59 ± 10.52         \\
        FML\cite{shenFederatedMutualLearning2023}                       & 82.22 ± 11.75                       & 88.29 ± 10.22                     & 91.07 ± 9.30                       & 92.38 ± 9.60          & {85.53 ± 13.11}         & 89.31 ± 11.89         & 91.07 ± 10.83         & 91.77 ± 9.76          \\ \hline
        FedL2T (Ours)                    & \textbf{87.33 ± 10.37}              & \textbf{92.06 ± 8.65}             & \textbf{95.21 ± 4.76}              & \textbf{97.09 ± 4.76} & {\textbf{89.79 ± 8.26}} & \textbf{93.94 ± 5.48} & \textbf{94.40 ± 4.95} & \textbf{95.72 ± 3.40} \\ \hline
        \end{tabular}
    }
\end{table*}

\subsection{Ablation Study}
To assess the contribution of each component within the proposed FedL2T framework, we conduct ablation studies on both the CHB-MIT and Renji datasets. Specifically, we evaluate two simplified variants: L2T-C, which utilizes only cross-client collaboration via a single peer teacher without any involvement from the central server, and L2T-CG, which incorporates both global and peer supervision but omits the proximal term used in the full FedL2T.

As shown in Table~\ref{tab: Fed-CHBMIT} and Table~\ref{tab: Fed-Private}, the cross-client variant L2T-C improves accuracy by 2.66\% and 3.51\% over the Local baseline on the CHB-MIT and Renji datasets, respectively. These gains already outperform several existing baselines, including FML, which also adopts teacher-based distillation. This highlights the effectiveness of cross-client collaboration over centralized global supervision in highly heterogeneous settings. Building upon L2T-C, the L2T-CG variant introduces global model guidance, forming a two-teacher framework with adaptive distillation strength control. This leads to performance improvements of 4.73\% and 5.17\% on the respective datasets, demonstrating the benefit of combining global and cross-client knowledge sources. Finally, by incorporating a proximal term into L2T-CG, the full FedL2T framework further stabilizes local updates, yielding the highest performance across both datasets. These results validate the necessity and complementarity of each component in the FedL2T design.

\subsection{Comparison of Convergence Rate}
We further analyze the convergence behaviors of all FL methods, including seven baselines and three ablation variants, as depicted in Fig.~\ref{fig:Fig3_curve}. FedL2T consistently achieves the highest accuracy in the final communication rounds on both datasets. Notably, it exhibits a relatively fast convergence rate, particularly in the early stage of training, second only to the L2T-CG variant. This rapid progress can be attributed the efficient escape from local optima enabled by cross-client collaboration. However, the dynamic nature of peer interactions in L2T-C and L2T-CG introduces larger fluctuations. Interestingly, Ditto demonstrates the most stable convergence owing to the proximal term that constrains local updates. Nevertheless, its performance declines in later stages due to its exclusive dependence on global model knowledge.

Building on these insights, the full FedL2T framework combines the stability of Ditto with the collaborative advantages of L2T-CG. As a result, it achieves a favorable balance between convergence speed, training stability, and final accuracy, reaching high performance with minimal oscillations.

\begin{table}[t!]
    \caption{Impact of $\lambda_c$ on Accuracy (\%) for Two Datasets.}
    \label{tab: HyperParameters}
    \renewcommand{\arraystretch}{1.05}
    \setlength{\tabcolsep}{5pt}  
    \resizebox{0.5\textwidth}{!}{ 
        \begin{tabular}{c|ccccccc}
            \hline
            $\lambda{c}$  & 0  & 0.25  & 0.50  & 0.75  & 1.00  & 1.25  & 1.50  \\ \hline
            CHBMIT   & 96.35 & 97.61 & \textbf{98.00} & 97.58 & 97.06 & 96.97 & 96.49 \\
            Renji  & 93.26 & 95.46 & \textbf{95.75} & 95.27 & 95.22 & 95.02 & 94.86 \\ \hline
        \end{tabular}
    }
\end{table}

\subsection{Evaluation under Limited Label Supervision}
To simulate practical scenarios with limited labeled data, we evaluate the proposed FedL2T framework against several representative FL methods on both the CHB-MIT and Renji EEG datasets. The experiments are conducted under varying proportions of labeled data (1\%, 5\%, 10\%, and 25\%) to assess performance under different levels of supervision.

As shown in Table~\ref{tab: Semi_Labels}, FedL2T consistently outperforms all baseline methods across both datasets and all labeling levels, demonstrating strong adaptability to limited supervision. Notably, when the labeled data ratio drops below 10\%, most existing methods fall short of the Local baseline. This performance degradation is particularly evident for methods such as FedMRL and Ditto, which, despite their advantages under fully supervised conditions, struggle to maintain effectiveness with reduced labeled data. The scarcity of labeled data may amplify inter-client heterogeneity, increasing variance among local models and hindering the quality of global aggregation. Consequently, methods relying heavily on global model guidance tend to be more vulnerable under such conditions. While FML also leverages global knowledge, it mitigates this issue through knowledge distillation, allowing more flexible and localized adaptation. Building on this idea, FedL2T further improves robustness by integrating global and cross-client supervision in a balanced manner and introducing adaptive weighting to ensure effective knowledge transfer under low-label settings. This design contributes to its consistently superior performance in low-resource federated learning scenarios.

\subsection{Analysis of Hyperparameter}
A key innovation of the proposed FedL2T framework lies in enabling each personalized model $\boldsymbol{P}_k$ to distill knowledge from two sources: a local transfer model $\boldsymbol{T}_k$ and a peer model $\boldsymbol{P}_c$. Each distillation path is adaptively weighted via task-loss-based normalization, reflecting the reliability of supervision. In addition, the hyperparameter $\lambda_c$ explicitly modulates the contribution from the peer model.

To assess the sensitivity of $\lambda_c$, we evaluate model performance on both datasets with $\lambda_c$ ranging from 0 to 1.5. As shown in Table~\ref{tab: HyperParameters}, accuracy increases with larger $\lambda_c$, reaching its peak at $\lambda_c = 0.5$, and declines thereafter. Specifically, the case of $\lambda_c = 0$ corresponds to a variant of FML augmented with adaptive distillation loss and proximal regularization. Its performance exceeds that of both FML and Ditto, demonstrating the value of these enhancements. Overall, the results validate the importance of balancing complementary knowledge sources, with $\lambda_c = 0.5$ serving as an effective default.

\section{Conclusions}
This paper presents FedL2T, a personalized federated learning framework with a two-teacher distillation strategy for seizure prediction. By allowing each client to learn from both a global model and a dynamically assigned peer, FedL2T facilitates more effective and diverse knowledge transfer. The framework further integrates adaptive multi-level distillation and proximal regularization to enhance training stability and personalization under non-IID conditions. Extensive experiments on both public and private EEG datasets demonstrate that FedL2T consistently outperforms state-of-the-art federated learning methods in terms of accuracy, convergence speed, and robustness under limited supervision. These results highlight the potential of FedL2T for deployment in real-world clinical settings and offer a promising direction for scalable personalized FL in healthcare. While the current framework focuses on homogeneous model architectures and random peer assignment for simplicity, it does not explicitly compare communication cost. However, the fast and stable convergence of FedL2T implicitly reduces the number of communication rounds. Future work will explore support for heterogeneous models, similarity-based peer selection, low-cost communication extensions, and broader biomedical applications.

\bibliographystyle{IEEEtran}

\begin{thebibliography}{10}
\providecommand{\url}[1]{#1}
\csname url@samestyle\endcsname
\providecommand{\newblock}{\relax}
\providecommand{\bibinfo}[2]{#2}
\providecommand{\BIBentrySTDinterwordspacing}{\spaceskip=0pt\relax}
\providecommand{\BIBentryALTinterwordstretchfactor}{4}
\providecommand{\BIBentryALTinterwordspacing}{\spaceskip=\fontdimen2\font plus
\BIBentryALTinterwordstretchfactor\fontdimen3\font minus \fontdimen4\font\relax}
\providecommand{\BIBforeignlanguage}[2]{{%
\expandafter\ifx\csname l@#1\endcsname\relax
\typeout{** WARNING: IEEEtran.bst: No hyphenation pattern has been}%
\typeout{** loaded for the language `#1'. Using the pattern for}%
\typeout{** the default language instead.}%
\else
\language=\csname l@#1\endcsname
\fi
#2}}
\providecommand{\BIBdecl}{\relax}
\BIBdecl

\bibitem{sunEmpoweringCrossPatientEpilepsy2024}
Y.~Sun, J.~Yu, and C.-T. Lu, ``Empowering {{Cross-Patient Epilepsy Diagnosis}} from {{Diverse-Sampling Low-Quality EEG Signals}},'' in \emph{2024 {{IEEE International Conference}} on {{Bioinformatics}} and {{Biomedicine}} ({{BIBM}})}, Dec. 2024, pp. 2460--2467.

\bibitem{segalPersonalizedPreictalEEG2025}
G.~Segal, N.~Keidar, M.~Herskovitz, and Y.~Yaniv, ``Personalized preictal {{EEG}} pattern characterization: Do timing and localization matter?'' \emph{Frontiers in Neuroscience}, vol.~19, May 2025.

\bibitem{liSpatioTemporalSpectralHierarchicalGraph2022}
Y.~Li, Y.~Liu, Y.-Z. Guo, X.-F. Liao, B.~Hu, and T.~Yu, ``Spatio-{{Temporal-Spectral Hierarchical Graph Convolutional Network With Semisupervised Active Learning}} for {{Patient-Specific Seizure Prediction}},'' \emph{IEEE Transactions on Cybernetics}, vol.~52, no.~11, pp. 12\,189--12\,204, Nov. 2022.

\bibitem{albrecht2016how}
J.~P. Albrecht, ``How the {{GDPR}} will change the world,'' \emph{European Data Protection Law Review}, vol.~2, no.~3, 2016.

\bibitem{McMahanMRHA17}
B.~McMahan, E.~Moore, D.~Ramage, S.~Hampson, and B.~A. y~Arcas, ``Communication-efficient learning of deep networks from decentralized data,'' in \emph{Proceedings of the 20th International Conference on Artificial Intelligence and Statistics (AISTATS)}, vol.~54, 2017, pp. 1273--1282.

\bibitem{zhang2025pfllib}
J.~Zhang, Y.~Liu, Y.~Hua, H.~Wang, T.~Song, Z.~Xue, R.~Ma, and J.~Cao, ``Pfllib: A beginner-friendly and comprehensive personalized federated learning library and benchmark,'' \emph{Journal of Machine Learning Research}, vol.~26, no.~50, pp. 1--10, 2025.

\bibitem{liDittoFairRobust2021}
T.~Li, S.~Hu, A.~Beirami, and V.~Smith, ``Ditto: {{Fair}} and {{Robust Federated Learning Through Personalization}},'' in \emph{Proceedings of the 38th {{International Conference}} on {{Machine Learning}} (ICML)}, Jul. 2021, pp. 6357--6368.

\bibitem{liFedBNFederatedLearning2020}
X.~Li, M.~Jiang, X.~Zhang, M.~Kamp, and Q.~Dou, ``{{FedBN}}: {{Federated Learning}} on {{Non-IID Features}} via {{Local Batch Normalization}},'' in \emph{International {{Conference}} on {{Learning Representations}} (ICLR)}, Oct. 2020.

\bibitem{collinsExploitingSharedRepresentations2021}
L.~Collins, H.~Hassani, A.~Mokhtari, and S.~Shakkottai, ``Exploiting {{Shared Representations}} for {{Personalized Federated Learning}},'' in \emph{Proceedings of the 38th {{International Conference}} on {{Machine Learning}} (ICML)}, Jul. 2021, pp. 2089--2099.

\bibitem{moraKnowledgeDistillationFederated2024}
A.~Mora, I.~Tenison, P.~Bellavista, and I.~Rish, ``Knowledge {{Distillation}} in {{Federated Learning}}: {{A Practical Guide}},'' in \emph{Proceedings of the 33th International Joint Conference on Artificial Intelligence (IJCAI)}, Aug. 2024.

\bibitem{shenFederatedMutualLearning2023}
T.~Shen, J.~Zhang, X.~Jia, F.~Zhang, Z.~Lv, K.~Kuang, C.~Wu, and F.~Wu, ``Federated mutual learning: A collaborative machine learning method for heterogeneous data, models, and objectives,'' \emph{Frontiers of Information Technology \& Electronic Engineering}, vol.~24, no.~10, pp. 1390--1402, Oct. 2023.

\bibitem{sastryGeneticAlgorithms2005}
K.~Sastry, D.~Goldberg, and G.~Kendall, ``Genetic {{Algorithms}},'' in \emph{Search {{Methodologies}}: {{Introductory Tutorials}} in {{Optimization}} and {{Decision Support Techniques}}}, 2005, pp. 97--125.

\bibitem{baghersalimiPersonalizedRealTimeFederated2022}
S.~Baghersalimi, T.~Teijeiro, D.~Atienza, and A.~Aminifar, ``Personalized {{Real-Time Federated Learning}} for {{Epileptic Seizure Detection}},'' \emph{IEEE Journal of Biomedical and Health Informatics}, vol.~26, no.~2, pp. 898--909, Feb. 2022.

\bibitem{saemaldahrPatientSpecificPreictalPatternAware2023a}
R.~Saemaldahr and M.~Ilyas, ``Patient-{{Specific Preictal Pattern-Aware Epileptic Seizure Prediction}} with {{Federated Learning}},'' \emph{Sensors}, vol.~23, no.~14, p. 6578, Jan. 2023.

\bibitem{FedESDFederatedLearning2023}
W.~Ding, M.~Abdel{-}Basset, H.~Hawash, S.~Abdel{-}Razek, and C.~Liu, ``Fed-{{ESD}}: {{Federated}} learning for efficient epileptic seizure detection in the fog-assisted internet of medical things,'' \emph{Information Sciences}, vol. 630, pp. 403--419, Jun. 2023.

\bibitem{baghersalimiDecentralizedFederatedLearning2024}
S.~Baghersalimi, T.~Teijeiro, A.~Aminifar, and D.~Atienza, ``Decentralized {{Federated Learning}} for {{Epileptic Seizures Detection}} in {{Low-Power Wearable Systems}},'' \emph{IEEE Transactions on Mobile Computing}, vol.~23, no.~5, pp. 6392--6407, May 2024.

\bibitem{dingFedMWADModulewiseWeighted2025}
Y.~Ding, W.~Zhao, and K.~Huang, ``{{FedMWAD}}: {{Module-wise}} weighted aggregation federated learning combined with {{Ditto}} for patient-independent seizure prediction,'' \emph{Information Processing \& Management}, vol.~62, no.~6, p. 104253, Nov. 2025.

\bibitem{romeroFitNetsHintsThin2015}
A.~Romero, N.~Ballas, S.~E. Kahou, A.~Chassang, and Y.~Bengio, ``{{FitNets}}: {{Hints}} for {{Thin Deep Nets}},'' in \emph{International {{Conference}} on {{Learning Representations}} (ICLR)}, 2015.

\bibitem{2020MiniLM}
W.~Wang, F.~Wei, L.~Dong, H.~Bao, N.~Yang, and M.~Zhou, ``Minilm: Deep self-attention distillation for task-agnostic compression of pre-trained transformers,'' in \emph{Advances in Neural Information Processing Systems (NeurIPS)}, vol.~33, 2020, pp. 5776--5788.

\bibitem{wuCommunicationefficientFederatedLearning2022}
C.~Wu, F.~Wu, L.~Lyu, Y.~Huang, and X.~Xie, ``Communication-efficient federated learning via knowledge distillation,'' \emph{Nature Communications}, vol.~13, no.~1, p. 2032, Apr. 2022.

\bibitem{zhangDeepMutualLearning2018a}
Y.~Zhang, T.~Xiang, T.~M. Hospedales, and H.~Lu, ``Deep {{Mutual Learning}},'' in \emph{2018 {{IEEE}}/{{CVF Conference}} on {{Computer Vision}} and {{Pattern Recognition (CVPR)}}}, Jun. 2018, pp. 4320--4328.

\bibitem{Shoeb2009ApplicationOM}
A.~H. Shoeb, ``Application of machine learning to epileptic seizure onset detection and treatment,'' Ph.D. dissertation, Harvard-MIT Health Sciences and Technology, Massachusetts Institute of Technology, 2009.

\bibitem{yi2024FedMRL}
L.~Yi, H.~Yu, C.~Ren, G.~Wang, X.~Liu, and X.~Li, ``Federated model heterogeneous matryoshka representation learning,'' in \emph{Advances in Neural Information Processing Systems (NeurIPS)}, 2024.

\end{thebibliography}

\end{document}